# Kernelized Weighted SUSAN based Fuzzy C-Means Clustering for Noisy Image Segmentation

Satrajit Mukherjee[1], Bodhisattwa Prasad Majumder [1], Aritran Piplai[2], and Swagatam Das[3]

[1]Dept. of Electronics and Telecomm. Engg., Jadavpur University, Kolkata 700 032, India.
[2]Dept. of Computer Science and Engg., Jadavpur University, Kolkata 700 032, India.
[3]Electronics and Communication Sciences Unit, Indian Statistical Institute, Kolkata – 700 108, India.
E-mails:satra0293@gmail.com, mbodhisattwa@gmail.com, aritran.piplai@gmail.com, swagatam.das@ieee.org,

*Abstract*—The paper proposes a novel Kernelized image segmentation scheme for noisy images that utilizes the concept of Smallest Univalue Segment Assimilating Nucleus (SUSAN) and incorporates spatial constraints by computing circular colour map induced weights. Fuzzy damping coefficients are obtained for each nucleus or center pixel on the basis of the corresponding weighted SUSAN area values, the weights being equal to the inverse of the number of horizontal and vertical moves required to reach a neighborhood pixel from the center pixel. These weights are used to vary the contributions of the different nuclei in the Kernel based framework. The paper also presents an edge quality metric obtained by fuzzy decision based edge candidate selection and final computation of the blurriness of the edges after their selection. The inability of existing algorithms to preserve edge information and structural details in their segmented maps necessitates the computation of the edge quality factor (EQF) for all the competing algorithms. Qualitative and quantitative analysis have been rendered with respect to state-of-the-art algorithms and for images ridden with varying types of noises. Speckle noise ridden SAR images and Rician noise ridden Magnetic Resonance Images have also been considered for evaluating the effectiveness of the proposed algorithm in extracting important segmentation information.

*Keywords*- *SUSAN*, *Circular color map, Edge Quality Factor, kernel, SAR, MRI, segmentation accuracy.*

## I. Introduction

Image segmentation [1] constitutes an important part of image processing which has various applications in the fields of feature extraction and object recognition. The goal of image segmentation methods is to cluster the pixels of an image into salient regions and hence these methods mainly involve various clustering techniques [2-6]. These clustering techniques separate a set of vectors or data points into different non-overlapping groups or regions such that each individual group or region, namely cluster, consists of similar kind of vectors or data points which are referred to as the members of that cluster. Recently researchers have proposed fuzzy segmentation methods which assign fuzzy membership values [7] to each image pixel according to its likelihood of belonging to various clusters. But, practically, in real-life problems, the digital image, to be segmented, is corrupted with various types of noises. Thus noisy image segmentation has become a challenge for classical segmentation methods because it requires both adequate removal of noise as well as preservation of the unique structural characteristics of the image like sharp edges, junctions and contours.

Fuzzy c-means (FCM) [8][9] clustering partitions a dataset or a set of image pixels, into *c* pre-defined number of clusters and assigns fuzzy membership values to each image pixel for its tendency to belong to a specific cluster. But this conventional method is not immune to noise and does not include spatial information in association with every individual pixel.

An enhanced FCM clustering method (EnFCM) [10] was proposed by Szilagyi *et al.*, on the basis of a linearly-weighted summed image formed by aggregating information from the local neighborhood of every pixel and original image. Cai *etal.* formulated a spatial similarity measure by utilizing both gray-level and spatial information to generate a non-linearly weighted image in the fast generalized FCM (FGFCM) [11] segmentation method. But the disadvantage of these methods is their dependency on several heuristic parameters which vary as the complexity of the digital image changes, hence leading to non-robustness. It is very difficult to choose these heuristics optimally, especially when the image is itself noise-ridden.

In order to eliminate the problem of excessive parameterization, Stelios *et al.* introduced a parameter-free fuzzy local information c-means clustering (FLICM) [12] method. Furthermore, a variant of this method, RFLICM [13], was introduced by Gong *et al.* but the method does not involve spatial constraints. Both these methods fail to accurately preserve the edge information in images as they produce blurry edges.

Most of the existing clustering schemes, including the above-mentioned methods, use Euclidean norm, which serves to be non-robust in case of non-Euclidean input data set. Kernel based methods [14]-[17] of segmentation transform data points; in this case, image features in the lower dimension inner product space to a higher dimensional space using non-linear mapping, thereby facilitating the segmentation process.

The existing kernel based image segmentation methods perform better segmentation of noisy images than classical segmentation methods; but they still suffer from their own drawbacks. For instance, the method proposed by Chen *et al.* [18] uses the mean of the surrounding pixels of a particular image pixel as a measure of spatial information. As a result of this, equal weights are assigned to all of the surrounding pixels of a particular pixel, which does not accurately convey the spatial contribution of different neighbors located at different distances from the pixel under consideration. More importantly, this method does not consider the gray-level or pixel intensity deviations in a particular neighborhood window around a pixel of concern.

Gong *e. al.* [19] recently proposed a kernel based fuzzy clustering scheme that takes into account both spatial constraints and neighborhood information. Their method proposed a trade-off weighted fuzzy factor that changes the contribution of neighborhood pixels in accordance with local coefficients of variation and independent noise distributions in localized square-shaped neighborhood windows. Our proposed method incorporates spatial constraints and local information by calculating the weighted mean of the surrounding pixels, the weights being dependent on circular color map [20] induced distances between the coordinates of the center pixel and that of the surrounding pixels. Circular color map induced weights have been used instead of Cartesian distance dependent ones so as to accurately portray the spatial damping for circularly shaped neighborhood masks. However, the foundation of our algorithm lies in extracting the weighted SUSAN [21][22] area values from all localized windows and forming a composite distribution of this weighted area over the entire image. Fuzzy non-homogeneity coefficients or damping coefficients are then derived by transforming the spatial domain localized weighted SUSAN area values into fuzzy domain values by utilizing the standard deviation of the composite distribution. The motivation for utilizing circular neighborhood masks and their corresponding SUSAN area information, instead of square neighborhood windows as used by Gong *et al.* in [19], is that the former has been used in various other image processing applications [22] to accurately preserve the information contained in edges, junctions and corners. To evaluate the effectiveness of the competing algorithms in preserving edge structure, we have devised a novel and accurate fuzzy decision based Edge Quality Factor (EQF) that incorporates the concepts of fuzzy rule based edge pixel estimation as discussed in [23] and a no-reference blur metric proposed in [24]. In the point of noise immunity, our method achieves more robustness than the other competing algorithms as shown by experimental results for different kinds of noise such as Salt and Pepper, Speckle, Gaussian, Poisson and Rician noise. Two speckle noise ridden Synthetic Aperture Radar (SAR) [25][26] images and two Rician [27][28] noise ridden medical image are considered for testing.

The organization of the paper is as follows:-
Section II provides the framework of the original kernel based work proposed by Chen *et al.* Section III introduces the weighted neighborhood information while sections IV and V present the need for computing weighted SUSAN area and fuzzy damping coefficients respectively. Section VI proposes the modified Kernel based objective function while Section VII provides experimental results. Applications to SAR and Medical Images and computational complexities are found in Sections VIII and IX while section X concludes the proceedings.

## II. FRAMEWORK OF THE ORIGINAL KERNEL BASED IMAGE SEGMENTATION

A spatial constraint based variant of FCM was proposed by Chen *et al.* in [18] whose objective function is given in Eq. (1) :-

$$J_m = \sum_{i=1}^{c}\sum_{k=1}^{N} u_{ik}^m \|x_k - v_i\|^2 + \frac{\alpha}{N_R}\sum_{i=1}^{c}\sum_{k=1}^{N} u_{ik}^m \sum_{r \in N_k}\|x_r - v_i\|^2 \quad (1)$$

The second part of the function in Eq. (1) stands for spatial information related to each image pixel, which eliminates the shortcomings of classical FCM. Though it tries to maintain homogeneity among neighborhood pixels, this method is burdened with a hefty computational overhead since all the pixels in a particular neighborhood window are needed to be considered in each iteration.

A simple modification has eliminated this problem and this was achieved by computing the term $\frac{1}{N_R}\sum_{r \in N_k}\|x_r - v_i\|^2$ as $\frac{1}{N_R}\sum_{r \in N_k}\|x_r - \overline{x_k}\|^2 + \|\overline{x_k} - v_i\|^2$, where $\overline{x_k}$ represents the mean of the surrounding pixels in a particular window. This modification takes less computational time as $\overline{x_k}$ can be calculated in advance. Hence the objective function boils down to the one presented in Eq. (2).

$$J_m = \sum_{i=1}^{c}\sum_{k=1}^{N} u_{ik}^m \|x_k - v_i\|^2 + \alpha \sum_{i=1}^{c}\sum_{k=1}^{N} u_{ik}^m \|\overline{x_k} - v_i\|^2 \quad (2)$$

Kernel-induced distances are used over this method by Chen *et al.* to improve the clustering scheme. A non-linear mapping $\Phi$ was introduced such as:- $\Phi: \mathbf{x} \in X \subseteq R^d \to \Phi(\mathbf{x}) \in F \subseteq R^H (d \ll H)$, which transforms a vector to a higher dimension. The mathematics involved in it, shows the transformation in Eq. (3):-

If $\mathbf{x} = [x_1, x_2]^T$ and $\Phi(\mathbf{x}) = [x_1^2, \sqrt{2}x_1x_2, x_2^2]^T$ then the inner product will be:-
$\Phi(\mathbf{x})^T\Phi(\mathbf{y}) = [x_1^2, \sqrt{2}x_1x_2, x_2^2]^T[y_1^2, \sqrt{2}y_1y_2, y_2^2] = (\mathbf{x}^T\mathbf{y})^2 = K(\mathbf{x}, \mathbf{y})$ (3)

This Kernel function $K(\mathbf{x}, \mathbf{y})$ is used to avoid the use of transformation matrix, ensuring an improvement in inner product.

$$K(\mathbf{x}, \mathbf{y}) = \exp\left(\frac{-(\sum_{i=1}^{d}|x_i - y_i|^a)^b}{\sigma^2}\right) \quad (4)$$

Eq. (4) provides a typical example of a Kernel function where *d* denotes the dimension of the vector and *a*>0; 1≤*b*≤2 and $\sigma$ is the variance of the Kernel function; $K(x, x) = 1$ for all *x*; whereas, a polynomial Kernel of degree *p* can be written as in Eq. (5)

$$K(\mathbf{x}, \mathbf{y}) = (\mathbf{x}^T\mathbf{y} + 1)^p \quad (5)$$

Kernel space can be constructed using Kernel functions instead of inner products. Centroids were taken in the original space instead of in a higher dimension for better interpretation of results. On the basis of these mathematical formulations, the objective function boiled down to the one in Eq. (6)

$$J_m = \sum_{i=1}^{c}\sum_{k=1}^{N} u_{ik}^m \|\Phi(\mathbf{x}_k) - \Phi(\mathbf{v}_i)\|^2 \quad (6)$$

Then a Kernelized substitution produced Eq. (7).

$$\|\Phi(\mathbf{x}_k) - \Phi(\mathbf{v}_i)\|^2 = (\Phi(\mathbf{x}_k) - \Phi(\mathbf{v}_i))^T(\Phi(\mathbf{x}_k) - \Phi(\mathbf{v}_i))$$
$$= \Phi(\mathbf{x}_k)^T\Phi(\mathbf{x}_k) - \Phi(\mathbf{x}_k)^T\Phi(\mathbf{v}_i) - \Phi(\mathbf{v}_i)^T\Phi(\mathbf{x}_k) + \Phi(\mathbf{v}_i)^T\Phi(\mathbf{v}_i)$$
$$= K(\mathbf{x}_k, \mathbf{x}_k) + K(\mathbf{v}_i, \mathbf{v}_i) - 2K(\mathbf{x}_k, \mathbf{v}_i) \quad (7)$$

Chen *et al.* finally proposed the original Kernel based objective function, as given in Eq. (8).

$$JS^\Phi = \sum_{i=1}^{c}\sum_{k=1}^{N}(u_{ik}^m(1 - K(x_k, v_i)) + \alpha \sum_{i=1}^{c}\sum_{k=1}^{N} u_{ik}^m(1 - K(\bar{x}_k, v_i)), \quad (8)$$

where the partition matrix values and centroids are presented as in Eqs. (9) and (10) respectively.

$$u_{ik} = \frac{\left((1 - K(x_k, v_i)) - \alpha(1 - K(\bar{x}_k, v_i))\right)^{-\frac{1}{m-1}}}{\sum_{j=1}^{c}\left((1 - K(x_k, v_i)) - \alpha(1 - K(\bar{x}_k, v_i))\right)^{-\frac{1}{m-1}}} \quad (9)$$

$$v_i = \frac{\sum_{k=1}^{n} u_{ik}^m (K(x_k,v_i)x_k + \alpha K(\bar{x}_k,v_i)\bar{x}_k)}{\sum_{k=1}^{n} u_{ik}^m (K(x_k,v_i) + \alpha K(\bar{x}_k,v_i))} \quad (10)$$

However the spatially varying contributions of the neighborhood were not taken into account. Hence, we have proposed certain spatial and neighborhood information based modifications of the original objective function that take into account fuzzy damping coefficients associated with each nucleus, derived using circular color map induced weighted SUSAN area values. The next section introduces the neighborhood mask shape and the circular color map induced weights.

### III. WEIGHTED NEIGHBOURHOOD INFORMATION

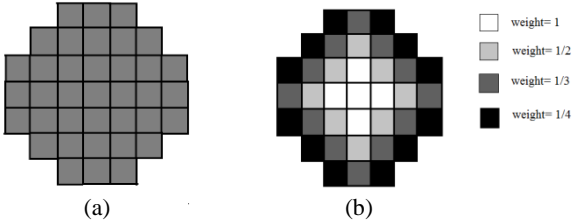

**Fig.1**: a) 37 pixels circular mask b) 37 pixels circular mask with circular color map induced weights

Most of the existing image segmentation algorithms fail to preserve the edges, junctions and contours present in the original noise-ridden image. The SUSAN edge detection algorithm [21][22] was introduced to achieve proper detection of junctions and contours in an image and this serves as a motivation to use a SUSAN area based circular mask to ensure the preservation of the edges and contours. For the computation of SUSAN area, a mask of 37 pixels, i.e. 36 pixels around a pixel of concern, is taken under consideration. The area spreads over 7 rows with the rows having 3,5,7,7,7,5,3 pixels respectively. The problem, however, lies in the fact that all the neighboring pixels in the entire mask are given equal importance or weights. To incorporate spatial information such that pixels have spatially varying contributions, circular color map induced weights are assigned to each and every pixel of the mask. The weight of a particular neighborhood pixel basically represents the inverse of the number of horizontal and vertical moves required to reach that pixel from the center pixel. Thus the entire circular mask is divided into 4 circular rings $G_1$, $G_2$, $G_3$ and $G_4$ with the contributions of the pixel members in the rings being 1, 1/2, 1/3 and 1/4 respectively as is indicated by Fig.1(b). The nucleus itself will have unit weight associated with it. Cartesian distances should not be used to determine the contributions of the neighbors since that will not reflect the actual circular nature of the mask. The members of the same circular ring will have different weights associated with them if Cartesian distances are used to determine the weights. For instance, the second most inner ring will have pixel members with both weights 1/2 as well as $1/\sqrt{2}$ associate with them. However, members belonging to the same ring must have same weights associated with them. Thus this circular color map induced weighted mean will be used in place of the arithmetic mean as an initial modification of the objective function proposed by Chen *et* al. The weights used in our approach are represented in Eq. (11).

$$w(r) = 1, if\ I(r)\epsilon\ G_1$$
$$= \frac{1}{2}, if\ I(r)\epsilon\ G_2$$
$$= \frac{1}{3}, if\ I(r)\epsilon\ G_3$$
$$= \frac{1}{4}, if\ I(r)\epsilon\ G_4, \quad \forall\ r\ \epsilon\ N_R \quad (11),$$

where $N_R$ is the circular neighborhood of the center pixel or the nucleus and $I(r)$ corresponds to any $rth$ pixel in neighborhood window $N_R$ including the nucleus.

This spatially and circularly varying weighted neighborhood information would be used to replace the arithmetic mean $\bar{x}_k$ with the circular colour map induced weighted mean $\bar{x}_{wk}$ which is computed as shown in Eq. (12):

$$\bar{x}_{wk} = \frac{\sum_{r\ \epsilon\ N_R}(w(r))*I(r)}{\sum_{r\ \epsilon\ N_R} w(r)}, \quad (12)$$

where $I(r)$ is the pixel intensity of a neighboring pixel $r\epsilon N_r$ and $w(r)$ is the circular pixel distance of the $r$-th neighbor from the center pixel or the neighbor. Thus an initial modification of the Kernel-based objective function can be given in Eq. (13):-

$$JS^\Phi = \sum_{i=1}^{c}\sum_{k=1}^{N}(u_{ik}^m(1-K(x_k,v_i)) + \alpha\sum_{i=1}^{c}\sum_{k=1}^{N}u_{ik}^m(1-K(\bar{x}_{wk},v_i)) \quad (13)$$

Here, we have not varied the contribution of the neighbors except for directly incorporating spatial constraints in the non-linear kernel mapping. The circular color mapped induced weights of neighbors around the nucleus i.e. $\bar{x}_{wk}$ have only been used to modify the inputs to the kernel mapping function in the second part of Eq. (13) i.e. $\alpha\sum_{i=1}^{c}\sum_{k=1}^{N}u_{ik}^m(1-K(\bar{x}_{wk},v_i))$ and have not been used explicitly as damping coefficients. The next subsections introduce fuzzy damping coefficients which would be used to further modify the Kernel based function by varying the contributions of every nucleus on the basis of weighted SUSAN area values computed for every circular neighborhood around the nuclei.

### IV. CIRCULAR COLOR MAP INDUCED WEIGHTED SUSAN AREA

The SUSAN area [21][22] is a metric for determining the number of neighbors that have similar intensity to the nucleus or the center pixel. The intensity of the nucleus is compared with all the surrounding pixels in the mask to compute the SUSAN area value. The deviations of the intensities of the 36 neighbors with respect to the intensity of nucleus are evaluated using Eq.(14).

$$\delta(r, r_0) = exp\left[-\left(\frac{(I(r)-I(r_o))}{t}\right)^6\right], \quad (14)$$

where '$r$' is the position of any neighborhood pixel, '$r_0$' is the position of the nucleus, $I(r)$ is the intensity of any pixel in the mask, $I(r_o)$ is the intensity of the nucleus and '$t$' is a parameter that determines the range of output of the equation.

The individual deviations for all the 36 neighbors computed by Eq. (14) are added to obtain the SUSAN area. Eq. (15) represents the SUSAN area.

$$D(r,r_o) = \sum_r \delta(r,r_o) \quad (15)$$

However, this sort of a calculation does not reflect the spatial information conveyed by the neighbors and thus the weights introduced in Section III are included in the individual deviation calculations to produce the modified deviations $\acute{\delta}(r, r_0)$ in Eq. (16),

$$\acute{\delta}(r, r_0) = w(r) * exp\left[-\left(\frac{I(r)-I(r_o)}{t}\right)^6\right] \quad (16)$$

where $w(r)$ can be computed from Eq.(11).
These individual deviations are then summed up using Eq. (17).

$$\acute{D}(r, r_o) = \sum_r \acute{\delta}(r, r_o) \quad (17)$$

As is evident from Eq. (16), if a neighboring pixel $I(r)$ has the same intensity as the nucleus, the output would be $w(r)$. A perfectly homogeneous region would have all the neighborhood pixel intensities equal to the nucleus intensity. In that case, the individual weighted deviations $\acute{\delta}(r, r_o)$ and the weighted sum of the outputs for all of the 36 neighboring pixels i.e. $\acute{D}(r, r_o)$ are given by Eqs, (18) and (19) respectively.

$$\acute{\delta}(r, r_o) = w(r) \quad \forall \ r \ \epsilon \ N_R \& \quad \forall I(r) = I(r_o) \quad (18)$$

$$\acute{D}(r, r_0) = \sum_r \acute{\delta}(r, r_o) = \sum_r w(r) = 16 \quad (19)$$

Thus the maximum value of the summed output or the weighted SUSAN area can be at the most $\sum_r w(r) = 16$ i.e. the sum of the circular colour map induced weights of all the pixels in $N_R$. However, that depends entirely on whether a perfectly homogeneous region of 37 pixels is present in the noise-ridden image. Thus, we choose to denote the maximum value of the weighted SUSAN area as calculated for a test image as $\acute{D}_{max}$. The choice of the parameter $t$ depends on the minimum value of the output of Eq. (17). The maximum intensity deviation $I(r) - I(r_o)$ can be 255 for a grayscale image and we will limit the minimum value of the Eq. (16) to 1/16 such that the minimum value of the summed output of Eq. (17) reduces to 1. Thus the value of the parameter $t$ can be obtained by solving the equation in Eq. (20).

$$exp\left[\left(\frac{-(255)}{t}\right)^2\right] = 1/16 \quad (20)$$

This yields the value of the parameter $t$ as 215.1424 such that the summed up output range of Eq. (17) i.e. the weighted SUSAN area lies within [1, 16].

## V. FUZZY DAMPING COEFFICIENTS

The initial weighted SUSAN area values proposed in Section IV are mapped to the fuzzy domain values [0, 1] using the Eq. (21) which represents a Gaussian membership [29]-[31].

$$\mu(\acute{D}) = exp\left(-\left(\frac{(\acute{D}_{max}-\acute{D})^2}{2*\sigma_{\acute{D}}^2}\right)\right) \quad (21)$$

where $\sigma_{\acute{D}}$ is the standard deviation of the values of all the spatial domain weighted SUSAN area values obtained for all the localized windows i.e. $\acute{D}$ and $\acute{D}_{max}$ is the maximum value of the measure globally obtained in an image. Thus computation of $\sigma_{\acute{D}}$ requires that the values of $\acute{D}$ for all the localized circular windows be recorded such that their standard deviation can be evaluated. The maximum value of $\acute{D}$ is '16' and the minimum value '1' as mentioned in section IV but it is dependent on the test image at hand.

The fuzzy mapping of the spatial domain non-homogeneity values increases the dynamic range of variation of the damping coefficients and associates fuzzy domain values in the range of [0, 1].

The entire Kernel based objective function can be thought of as a summation of the contribution from the nucleus and the contribution of its neighborhood. In case of a perfectly homogeneous region, the contributions of the neighboring pixels have to be taken into account and thus the contribution of the nucleus can be suppressed. With increase in non-homogeneity, the contribution of the nucleus in the objective function is increased. Higher membership values $\mu(\acute{D})$ correspond to more homogeneity and hence the damping coefficients required to decrease the contribution of the nucleus with increasing homogeneity is given by $s(k)$ for every $kth$ pixel in Eq. (22).

$$s(k) = 1 - \mu(\acute{D}) \quad (22)$$

where $\acute{D}$ is the weighted SUSAN area value associated with the $kth$ nucleus.

Pertaining to the problem of noise removal with preservation of proper edge and contour information, this modification of the SUSAN principle serves as a better measure of spatial information than taking Cartesian distance induced weights. We conducted our experiments with Cartesian induced weights too and also without taking any spatial constraints or spatially varying weights into account. Fig. 2(a)-(c) compare the segmentation maps produced by our proposed method i.e. KWSFCM with respect to those obtained by both no spatial constraint as well as Cartesian distance induced weights. As expected, Fig. 2(a) shows blurry edges since no spatial constraint was taken into consideration. Fig. 2(b) generated with Cartesian distance induced weights fail to suppress noise sufficiently due to the different contributions of pixel members belonging to the same circular ring in the circular mask while Fig. 2(c) obtained by KWSFCM shows sufficient removal of noise as well as preservation of accurate edge information.

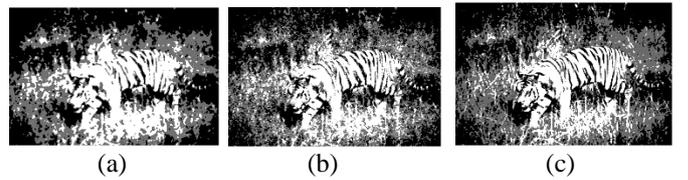

(a)            (b)            (c)

Fig. 2: a) Segmented image using original SUSAN mask  b) using Cartesian distance induced weights   c) using circular color map induced weights.

## VI. MODIFICATION OF OBJECTIVE FUNCTION

The final modified function incorporates both spatial constraints by using the circular colour map induced weighted pixel intensities as input to the Kernel map as well as non-homogeneity information by using the fuzzified damping coefficients $s(k)$ which increase the contribution of the nucleus with increasing non-homogeneity. The modified Kernel based equation can be presented in Eq. (23) as:

$$JS^\Phi = \sum_{i=1}^{c}\sum_{k=1}^{N}(s(k)*u_{ik}^m(1-K(x_k,v_i)) + \alpha\sum_{i=1}^{c}\sum_{k=1}^{N}u_{ik}^m(1-K(\bar{x}_{wk},v_i)) \quad (23)$$

where $s(k)$ is the damping coefficient evaluated for any $k$-th pixel, in accordance with Eq. (22).

Similarly, the partition matrix values $u_{ik}$ and the centroids $v_i$ are modified in Eqs. (24) and (25) respectively by incorporating the weighted mean and the suppressing coefficients. The values of the parameters $m$, $\alpha$ and $\sigma$ of the kernel have been taken as 2, 3.8 and 150 respectively as proposed by Chen *et* al. as the variations of these parameters do not significantly retard the performance of our algorithm.

$$u_{ik} = \frac{\left(s(k)*(1-K(x_k,v_i)) - \alpha(1-K(\bar{x}_{wk},v_i))\right)^{-\frac{1}{m-1}}}{\sum_{j=1}^{c}\left(s(k)*(1-K(x_k,v_i)) - \alpha(1-K(\bar{x}_{wk},v_i))\right)^{-\frac{1}{m-1}}} \quad (24)$$

$$v_i = \frac{\sum_{k=1}^{n}u_{ik}^m(s(k)*K(x_k,v_i)x_k + \alpha K(\bar{x}_{wk},v_i)\bar{x}_k)}{\sum_{k=1}^{n}u_{ik}^m(s(k)*K(x_k,v_i) + \alpha K(\bar{x}_{wk},v_i))} \quad (25)$$

The entire pseudocode of the algorithm is presented here. The optimization of the objective function is simply done using successive iteration method which is present in the pseudocode, showing necessary termination criterion for the optimization.

---

*PseudoCode of KWSFCM*

**Step 1)** Define the number of desired clusters c and Choose *c* prototype centroids of these clusters and set ε=0.001.

**Step 2)** Compute fuzzy damping coefficients to set up mathematical expressions for the modified objective function, partition matrix values and centroids.

**Step 3)** Update the partition matrix values using Eq (24)

**Step 4)** Update the centroids using Eq (25)

Repeat Steps 3)-4) until the following termination criterion is satisfied:

‖V$_{new}$- V$_{old}$‖ <ε

where V has been defined previously and ε has been introduced in step 1.

---

## VII. EXPERIMENTAL RESULTS

Experiments have been carried out on the test images taken from the Berkeley Segmentation Dataset-500 (BSDS-500) [http://www.eecs.berkeley.edu/Research/Projects/CS/vision/bsds)]. Images having different complexities and different distinguishing patterns have been taken to compare our results with those of other competing algorithms. Furthermore, a synthetic image has been used to determine the computational time of our proposed approach i.e. KWSFCM and to compare it with that of the existing methods. The size of the test images, which are taken from BSDS is 481x321. The size of the synthetic image was varied from 100x100 to 600x600 to generate the plot for computational complexities of all competing algorithms.

### A. Qualitative Analysis

Qualitative analysis has been rendered with respect to three test images with varying complexities. NNCut algorithm [32], one of the competing algorithms, is basically a *Nystrom* method based spectral graph grouping algorithm whereas FLICM, RFLICM, WFLICM and KWFLICM are the other state-of-the-art noisy image segmentation algorithms.

The original images without noise are in Fig. 3.

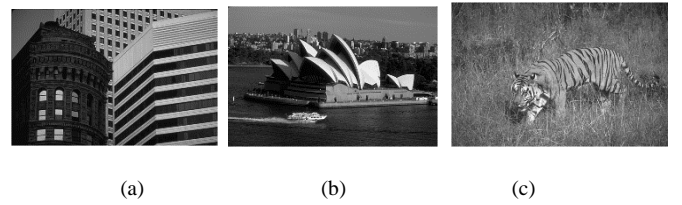

(a)        (b)        (c)

**Fig. 3**: a) House b) Sydney c) Tiger Images

The analysis can be done qualitatively on the basis of Figs. 4, 5 and 6 where Figs. 4(a), 5(a), 6(a) are the original image ridden with noise. Precisely, Fig. 4a) represents a 30% Salt & Pepper noise added image of two buildings or houses, 5(a) represents a 30% Gaussian noise added image of the Sydney house while 6(a) represents a Poisson noise added image of a tiger. Poisson noise cannot be artificially added. It is generated from the image data itself. 3-level segmentation has been rendered for these test images.

Qualitative analysis shows that the segmented images obtained using NNCut algorithm in Figs. 4b), 5b, 6b still contain an appreciable amount of noise as can be seen from speckles left. However, it does manage to preserve the structural details of the image. The main disadvantage of FLICM and RFLICM algorithms, as can be shown from Figs. 4(c)-(d), 5(c)-(d) and 6(c)-(d) is that these methods are associated with blurry edges and distorted image structures though they remove noise selectively. The WFLICM and KWFLICM methods show particularly good results in case of salt and pepper noise but fail to maintain their quality of performance in case of distributed noise like Gaussian and Poisson as it is evident from the Figs. 5(e)-(f) and 6(e)-(f). KWSFCM not only removes all type of noise but also conserves the shapes of different image structures and sharp edges present in the image. A detailed qualitative analysis easily shows the superiority and robustness of our algorithm to various type of noise.

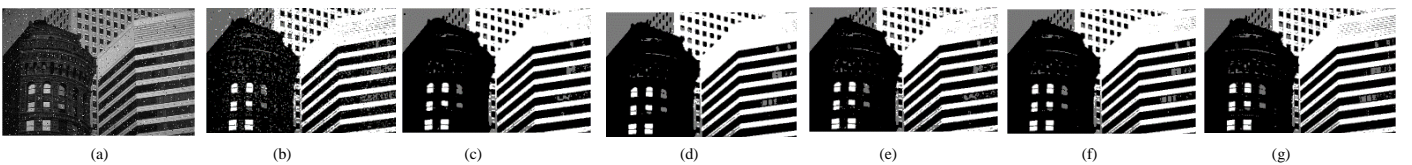

(a)    (b)    (c)    (d)    (e)    (f)    (g)

**Fig. 4**: a) Salt & pepper noise (30%) added House b) NNCUT c) FLICM d) RFLICM e) WFLICM f) KWFLICM g) KWSFCM

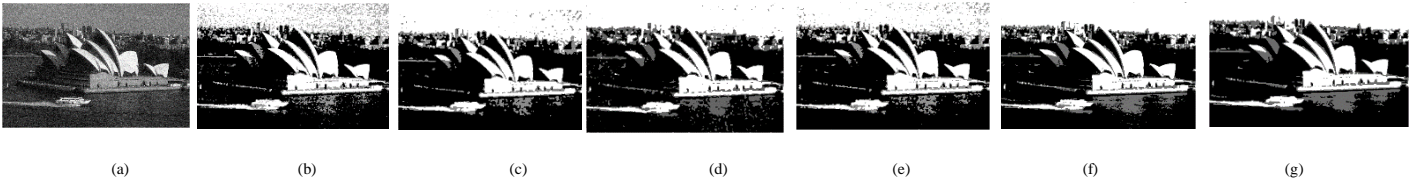

**Fig. 5**: a) Gaussian noise (30%) added Sydney b) NNCUT c) FLICM d) RFLICM e) WFLICM f) KWFLICM g) KWSFCM

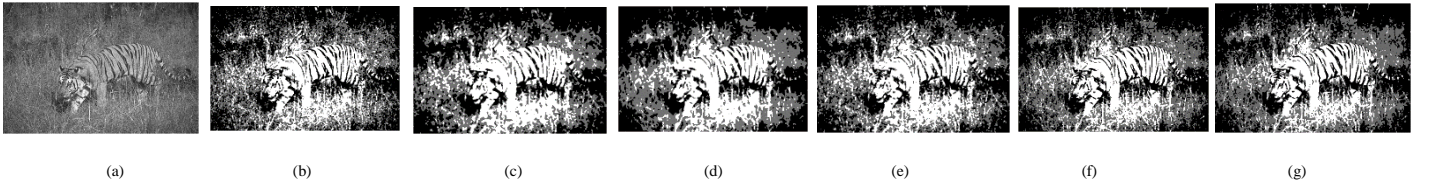

**Fig. 6**: a) Poisson noise added Tiger b) NNCUT c) FLICM d) RFLICM e) WFLICM f) KWFLICM g) KWSFCM

## B. QUANTITATIVE MEASURES

We examined the abovementioned test images quantitatively on the basis of the metrics discussed in this sub-section. To ensure the robustness of our algorithm, we varied the amount of Salt & Pepper noise and Gaussian noise between 20% and 30%. Poisson noise is generated from the image data itself instead of being superficially added. 25 independent runs for all test images were taken to average the results and then the comparison with other competing algorithms was made. Best results have been marked in bold face.

### 1) Measure dependent on ground truth

Segmentation Accuracy (SA) [33] is considered an important segmentation metric as it determines the fraction of correctly assigned pixels to a particular cluster, hence giving us a clear idea about the de-noising capabilities of different algorithms used in our experiments. This SA can be defined as the sum of the pixels which are correctly assigned to a particular cluster divided by the sum of the total number of pixels. The mathematical form can be written as in Eq. 26.

$$SA = \sum_{i=1}^{c} \frac{A_i \cap R_i}{\sum_{j=1}^{c} R_j} \quad (26)$$

Here $c$ is the number of clusters, $A_i$ is the set of pixels which forms the $i$-th cluster as per the algorithm and $R_i$ represents the referenced image's set of pixel which forms its i-th cluster.

The reference or ground images were generated by applying the classical FCM method without adding any noise to the images and then segmentation accuracy was calculated for each noise-ridden image with respect to these ground truth images.

Table I depicts the maximum Segmentation Accuracy of our proposed method with respect to all test images for all noise types of varying concentrations as compared to the competing algorithms. Higher value of SA indicates more appropriate clustering.

The pixels of the noisy image need to be assigned to those clusters which would have been assigned to the pixels had there been no noise in the image. Our algorithm adequately removes noise and assigns the pixels to proper clusters as is indicated by the maximum values of SA recorded in Table I. The NNCut algorithm fails to adequately remove noise, as a result of which many pixels have been assigned to inappropriate clusters. Thus it has the lowest values of SA associated with it. A qualitative look at Figs 4(b), 5(b) and 6(b) show the inability of the NNCut algorithm to remove noise as can be seen from the speckles in the images that have been assigned to different clusters with respect to their immediate background. Similarly, the lower values of SA for the other algorithms can be attributed to their insufficient removal of noise with respect to our algorithm. In addition, the FLICM and RFLICM algorithms produce blurry edges which indicate that the edge or contour pixels have been assigned to improper clusters, a problem which is eradicated completely by KWSFCM.

### 2) Measure independent of ground truth

In the absence of absolute ground truth images, a quantitative comparison on the basis of Segmentation Accuracy is impossible. Hence we have used a ground truth independent measure which is basically an entropy based objective function [34] whose minimization ensures that the similarity between the intra cluster pixels is maximized and similarity between pixels residing in different regions is minimized. Eq. (27) defines the region based entropy measure as:-

$$H(R_j) = -\sum_{m \in V_j} \frac{L_j(m)}{S_j} \log \frac{L_j(m)}{S_j} \quad (27)$$

where $R_j$ denotes the region of the image which makes up the $j$th cluster. $L_j(m)$ denotes the number of pixels in the region $R_j$ which have gray level values of 'm'. $V_j$ is the set of all pixel intensities that are present in the region $R_j$.

Cardinality is denoted by $S_j=|R_j|$ which also signifies the number of pixels in the region $R_j$ region. The region entropy for segmented image can be formulated as in Eq. (28)

$$H_r(I) = \sum_{j=1}^{C} \left(\frac{S_j}{S_I}\right) H(R_j) \quad (28)$$

Moreover, the entropy for the layout is defined in Eq. (29) as:

$$H_l(I) = -\sum_{j=1}^{C} \left(\frac{S_j}{S_I}\right) \log \left(\frac{S_j}{S_I}\right) \quad (29)$$

A final entropy based objective function can be derived combining both of the abovementioned entropies and can be formulated as in Eq. (30):-

$$E = H_l(I) + H_r(I) \quad (30)$$

Lower value of $E$ indicates superior clustering scheme. Table II shows minimum $E$ with respect to three test images with different noise types and for all competing algorithms. The Salt & Pepper noise added House image has been taken to represent a standard Salt & Pepper noise added image while the noisy images of Sydney and Tiger represent Gaussian noise added and Poisson noise added images respectively. Lower the value of $E$, the better is the clustering of pixels. Our algorithm achieves lowest values of $E$ which indicates optimal immunity to noise and outliers.

Here, we present an iterative convergence of the cluster sets for the Salt & Pepper added House Image as can be seen from Fig. 7 which depicts the change in partition matrix values noted at $1^{st}$ (u1), $5^{th}$(u2), $10^{th}$(u3) and at the last iteration(u) i.e. $22^{nd}$ (in this case) for which the error becomes less than ε. The curves of u1, u2, u3 and u are present in Fig. 7. Also, the iterative changes of the centroids i.e. V1 ($1^{st}$ iteration), V2 ($5^{th}$ iteration), V3 ($10^{th}$ iteration) and V ($22^{nd}$ iteration) are noted and plotted in Fig. 8. Due to space constraint, iterative changes of partition matrix values and centroids for other test images have been served in supplementary file.

**Table I:** Segmentation Accuracy (SA%) for all test images for all competing algorithms

| Noise | Image | NN Cut | FLICM | RFLICM | WFLICM | KWFLICM | Proposed method |
|---|---|---|---|---|---|---|---|
| 20% Salt & Pepper | House | 96.4802 | 99.5982 | 99.7098 | 99.7977 | 99.8189 | **99.9184** |
| 30% Salt & Pepper | | 94.0541 | 99.4439 | 99.6145 | 99.6457 | 99.7234 | **99.8356** |
| 20%Gaussian | | 92.9085 | 99.0375 | 99.3109 | 99.7002 | 99.7234 | **99.9078** |
| 30%Gaussian | | 89.0501 | 98.7341 | 98.8995 | 99.1349 | 99.6020 | **99.8095** |
| Poisson | | 95.0784 | 97.8134 | 98.9976 | 99.1295 | 99.8098 | **99.9005** |
| 20% Salt & Pepper | Sydney | 95.2405 | 99.1207 | 99.4021 | 99.6234 | 99.8451 | **99.9256** |
| 30% Salt & Pepper | | 92.0631 | 99.2016 | 99.4291 | 99.6192 | 99.6854 | **99.7984** |
| 20%Gaussian | | 91.8996 | 99.4501 | 99.4697 | 99.6901 | 99.7255 | **99.8540** |
| 30%Gaussian | | 87.4595 | 99.4289 | 99.5007 | 99.6874 | 99.7106 | **99.7998** |
| Poisson | | 92.9858 | 97.4110 | 98.8851 | 99.6781 | 99.8562 | **99.9259** |
| 20% Salt & Pepper | Tiger | 95.5667 | 99.4104 | 99.4747 | 99.6891 | 99.7375 | **99.9004** |
| 30% Salt & Pepper | | 93.0673 | 99.2992 | 99.5893 | 99.6651 | 99.7130 | **99.8812** |
| 20%Gaussian | | 92.0076 | 99.1108 | 99.2154 | 99.6870 | 99.7201 | **99.8997** |
| 30%Gaussian | | 88.1398 | 98.8921 | 99.2075 | 99.4409 | 99.5432 | **99.8092** |
| Poisson | | 94.1207 | 98.2118 | 98.8956 | 99.3401 | 99.8264 | **99.9103** |

**Table II:** Entropy measure for all test images for all competing algorithms

| Image with noise | Metric | NN Cut | FLICM | RFLICM | WFLICM | KWFLICM | Proposed method |
|---|---|---|---|---|---|---|---|
| House (30% Salt & Pepper) | $H_r(L)$ | 1.9071 | 1.8957 | 1.8860 | 1.8795 | 1.8134 | 1.8067 |
| | $H_l(L)$ | 1.1018 | 0.4031 | 0.3825 | 0.3690 | 0.4167 | 0.3594 |
| | $E$ | 3.0089 | 2.2988 | 2.2685 | 2.2485 | 2.2301 | **2.1661** |
| House (30% Gaussian) | $H_r(L)$ | 2.1632 | 1.9784 | 1.5568 | 1.5281 | 1.5127 | 1.5046 |
| | $H_l(L)$ | 0.8181 | 0.7430 | 0.7879 | 0.7050 | 0.6804 | 0.4835 |
| | $E$ | 2.9813 | 2.7214 | 2.3447 | 2.2331 | 2.1931 | **1.9881** |
| House (Poisson) | $H_r(L)$ | 1.7419 | 1.6744 | 1.5466 | 1.4121 | 1.3176 | 1.2144 |
| | $H_l(L)$ | 0.2568 | 0.3220 | 0.4410 | 0.5698 | 0.6610 | 0.5790 |
| | $E$ | 1.9987 | 1.9964 | 1.9876 | 1.9819 | 1.9786 | **1.7934** |
| Sydney (30% Salt & Pepper) | $H_r(L)$ | 2.5455 | 2.4264 | 2.1346 | 1.9917 | 1.8925 | 1.8123 |
| | $H_l(L)$ | 0.4230 | 0.4490 | 0.1452 | 0.2468 | 0.3264 | 0.3867 |
| | $E$ | 2.9685 | 2.8754 | 2.2798 | 2.2385 | 2.2189 | **2.1990** |
| Sydney (30% Gaussian) | $H_r(L)$ | 2.1932 | 2.0073 | 1.5807 | 1.5506 | 1.5345 | 1.5246 |
| | $H_l(L)$ | 0.8044 | 0.8937 | 0.8065 | 0.6998 | 0.6400 | 0.4739 |
| | $E$ | 2.9976 | 2.9012 | 2.3872 | 2.2504 | 2.1745 | **1.9985** |
| Sydney (Poisson) | $H_r(L)$ | 1.9866 | 1.8732 | 1.7823 | 1.7638 | 1.5954 | 1.5645 |
| | $H_l(L)$ | 0.4106 | 0.4130 | 0.3174 | 0.3345 | 0.3696 | 0.2948 |
| | $E$ | 2.3972 | 2.2862 | 2.0997 | 2.0983 | 1.9650 | **1.8593** |
| Tiger (30% Salt & Pepper) | $H_r(L)$ | 2.2669 | 2.1953 | 1.9038 | 1.7628 | 1.6747 | 1.6027 |
| | $H_l(L)$ | 0.5185 | 0.1156 | 0.4043 | 0.5269 | 0.5037 | 0.4940 |
| | $E$ | 2.7854 | 2.3109 | 2.3081 | 2.2897 | 2.1784 | **2.0967** |
| Tiger (30% Gaussian) | $H_r(L)$ | 2.5756 | 2.4550 | 2.1591 | 2.0145 | 1.9143 | 1.8330 |
| | $H_l(L)$ | 0.4276 | 0.4145 | 0.2947 | 0.2640 | 0.2623 | 0.2401 |
| | $E$ | 3.0032 | 2.8695 | 2.4538 | 2.2785 | 2.1766 | **2.0731** |
| Tiger (Poisson) | $H_r(L)$ | 2.1773 | 2.0719 | 1.7867 | 1.6586 | 1.5623 | 1.5014 |
| | $H_l(L)$ | 0.0314 | 0.0549 | 0.3009 | 0.3399 | 0.4254 | 0.3645 |
| | $E$ | 2.2087 | 2.1268 | 2.0876 | 1.9985 | 1.9877 | **1.8659** |

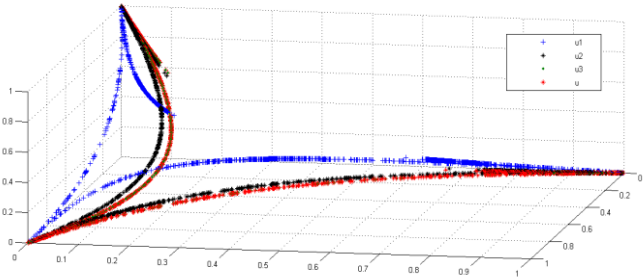

**Fig 7:** Iterative changes of partition matrix values

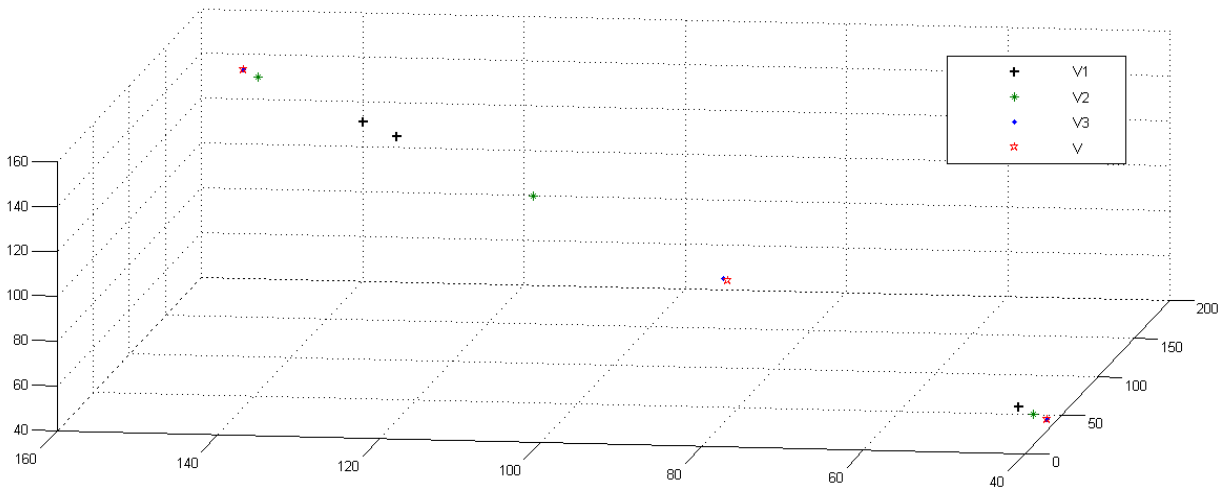

**Fig 8:** Iterative changes of centroid values

### 3) No-reference Fuzzy Rule based Edge Quality measure

A problem with most of the segmentation algorithms when applied to noise-ridden images is that they fail to preserve the quality of image structure in the form of edges, contours and junctions. Thus it becomes necessary to assess the quality of edges in the segmentation maps generated by the competing algorithms. In our work, we propose a no reference metric for assessing the quality of edges and quantifying the amount of blur introduced by blurry edges. The evaluation of this metric starts with a fuzzy rule based decision mechanism, for selecting edge candidates, that is motivated by the noise and image structure demarcation process used in a fuzzy image filtering algorithm proposed by [23]. After the decision process, the blur content in edges is evaluated by modifying the scheme for evaluation of blur ratio as proposed by Min Goo Choi *et al.* in [24].

#### a. Fuzzy Rule Based Decision for Edge Candidates

The decision process used in the method proposed in [24] takes into account only the horizontal and vertical derivatives for every pixel of concern. But our metric takes into account fuzzy derivative values along 8 directions given by the set $dir$ ={NW, W, SW, S, SE, E, NE, N} in order to correctly identify edge candidates that may be oriented along any of the 8 edge directions and not just along the horizontal or vertical direction.

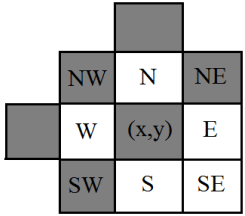

**Fig. 9:** 3x3 mask around the center pixel (x,y) and the pixels in gray are used to compute fuzzy derivative along the NW direction.

Each of the 8 fuzzy derivatives, along the 8 specified directions shown in Fig. 9, can be represented as a set of three derivatives. For example, the fuzzy derivative $\nabla^F_{d\epsilon\ dir}(x,y)$ for any $d\epsilon\ dir$ consists of three derivatives given by the set $\{\nabla^A_{d\epsilon\ dir}, \nabla^B_{d\epsilon\ dir}, \nabla^C_{d\epsilon\ dir}\}$. A detail of the pixel sets involved for computing the fuzzy derivative for each direction is provided in [23] and is also added in the supplementary file. An edge in an image is associated with large derivative values compared to homogeneous regions and noise and thus it is safe to discard a pixel as a non-edge candidate if at least 2 out of the 3 derivatives along any of the 8 directions are small. A parameter $K$ is used to determine whether the value of a derivative is small or large. The decision rule for the large membership function is given as in Eq. (31):-

$$\nabla^F_{d\epsilon\ dir}(x,y) \in m_k(u) \text{ if } \nabla^A_{d\epsilon\ dir} > K \qquad (31)$$

where $K$ was derived in [24] as shown in Eq. (32).
$$K = \alpha(1-\mu)\gamma_{N^2} \qquad (32)$$
μ is the expected value of all homogeneity values calculated around neighborhoods of sizes *NxN*. The individual $\mu$ calculations or $\mu_w$ have been done in accordance with Eq. (33),

$$\mu_w = 1 - \frac{W_{max} - W_{min}}{L} \qquad (33)$$

where $W_{max}$ and $W_{min}$ are the maximum and minimum pixel intensities in an *NxN* neighborhood of concern. Here *N* was taken to be 9 and the values of $\gamma_{N^2}$ were taken as presented in [23].

The final decision rule for an edge candidate is given as in Eq. (34):-
If
($\nabla^A_{d\epsilon\ dir}$ is large and $\nabla^B_{d\epsilon\ dir}$ is large)
Or
If ($\nabla^B_{d\epsilon\ dir}$ is large and $\nabla^C_{d\epsilon\ dir}$ is large)
Or
If ($\nabla^A_{d\epsilon\ dir}$ is large and $\nabla^C_{d\epsilon\ dir}$ is large) (34)

Then $C(x,y) = I(x,y)$,
i.e. in other words, a pixel $I(x,y)$ is considered as an edge candidate $C(x,y)$ if there are at least 2 derivatives out of 3 along any direction which belong to the large membership function.

#### b. Final Selection Of Edge Pixels

A final decision rule for the edge candidate is taken on the basis of 3-pixel wide derivatives calculated along the horizontal, vertical and diagonal directions that cover all possible orientations of an edge with respect to the center pixel concerned. This reduces some of the false positive edge candidates that may appear from the previous decision process. Eq. (35) provides the final decision rule. These derivative take into account the intensities of every pair of neighbors and thus the 8 dimensions mentioned before need not be considered for computing the Edge Quality Factor. They are required only for the edge candidate selection stage.

$$E(x,y) = 1 \text{ if } C(x,y) > \min\{C(x_{d\acute{i}r}, y_{d\acute{i}r})\}, \qquad (35)$$

$d\acute{i}r = \{h, v, d1, d2\}$ corresponds to horizontal, vertical and the two diagonal edge directions of the mask where $(x, y) \in n_r$ and $n_r$ is the 3x3 neighborhood around any pixel of concern.
Eq. (35) implies that an edge pixel will have greater intensity than its blurry neighbors.

#### c. Calculation of Inverse Blurriness

A measure called inverse blurriness was introduced in [24] but it only covered 3 pixel wide derivatives along horizontal and vertical directions. We have taken the two diagonals into consideration as well and computed 3 pixel wide derivatives along these two directions. The four derivatives along the horizontal, vertical directions and the diagonals whose set is given by $d\acute{i}r = \{h, v, d1, d2\}$, are presented in Eq. (36).

$$\begin{aligned}\nabla_h(x,y) &= |f(x, y+1) - f(x, y-1)| \\ \nabla_v(x,y) &= |f(x+1, y) - f(x-1, y)| \\ \nabla_{d1}(x,y) &= |f(x+1, y-1) - f(x-1, y+1)| \\ \nabla_{d2}(x,y) &= |f(x+1, y+1) - f(x-1, y-1)|\end{aligned}$$
(36)

The inverse blurriness values for the four directions are computed as in Eq. (37):-

$$BR_{d\epsilon dir}(x,y) = \frac{\left|f(x,y)-\frac{1}{2}*\nabla_{d\epsilon dir}(x,y)\right|}{\frac{1}{2}*\nabla_{d\epsilon dir}(x,y)} \quad (37)$$

*d. Decision rule for Blurred Edge*

The edge is considered blurred if the maximum of the Inverse Blurriness values for a pixel I(x,y) is less than a certain Threshold (*Th*) which was kept as 0.1 in the original work. The choice is prudent for our approach as well and the decision rule is presented in Eq. (38).

$$Blur(x,y) = \begin{cases} 1 & if \ \max(BR_{d\epsilon dir}(x,y) < Th \\ 0, & otherwise \end{cases} \quad (38)$$

*e. Computation of Edge Quality Factor*

A metric for quantifying the blurredness of edges is given by Eq. (39).

$$Blur\ ratio = Blur\_count/\ Edge\_count \quad (39)$$

where $Blur\_count$ is the number of blurry edges and $Edge\_count$ is the number of edge candidates determined by the fuzzy rule based mechanism.

Edge Quality Factor ($EQF$) defined in Eq. (40) assesses the quality of edges in the segmentation map. Lower the *Blur ratio*, higher is the EQF.

$$EQF = 1 - Blur\ ratio \quad (40)$$

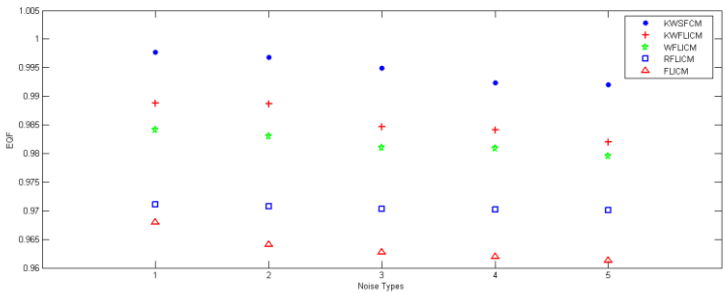

**Fig. 10**: EQF value obtained for competing algorithms for five types of noise as 20% Salt & Pepper, Poisson, 30%Salt & Pepper, 20% Gaussian and 30% Gaussian respectively.

NNCut algorithm fails to preserve noise while still preserving edge information. This algorithm has not been considered for evaluating EQF since the analysis of this factor should be done for algorithms that actively remove noise but selectively preserve edge information. Fig. 10 shows the values of $EQF$ for the remaining competing algorithms averaged over 20 benchmark images from BSDS, for the five types of noises. The *x* axis presents the five types of noises as 20% Salt & Pepper, Poisson, 30%Salt & Pepper, 20% Gaussian and 30% Gaussian respectively. Highest values of EQF are obtained by our algorithm for all sorts of noises, indicating that it has sufficiently preserved edge information while still managing to remove noise to a considerable extent.

### C. INCREASING THE NUMBER OF CLUSTERS

This clustering method is mainly based on spatial illumination deviations in the digital image. Based on this illumination diversity over the image, it is desirable to choose more number of clusters into which the test digital image has to be segmented. Choosing more number of cluster exposes more intricate details which can help in minute object detection. To show the effect, we choose a diversely illuminated image 'Hill' from BSDS-500, which contains differently illuminated layers as can be seen from the mountain region in the image and a 5-level clustering was applied to extract the intrinsic details present in the image. Fig. 11(a) presents the test image 'Hill', corrupted by noise. Fig 11(b) and 11(c) shows the segmented images with 3 level and 5 level clustering respectively. A close inspection of these images reveals that the distant layers of the mountain are not visible in the 3-level segmented image whereas the intrinsic details of those distant layers of the mountain can be clearly spotted in the 5-level segmented image.

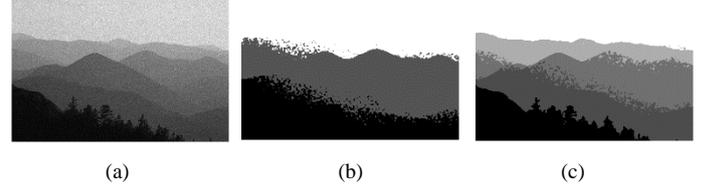

(a) (b) (c)

**Fig. 11**: a) Original Noisy Image 'Hill' b)3-level c)5-level segmentation using KWSFCM

### D. EXTENSION TO COLOR IMAGES

Every color image can be visualized as a combination of three primary components- Red, Green and Blue images. Each component can be considered as a gray-scale image and can be segmented in presence of noise. After segmentation, the three components can be concatenated which leads to a segmented color image as can be seen from Fig. 12b) while the noisy test color image is presented in Fig. 12a).

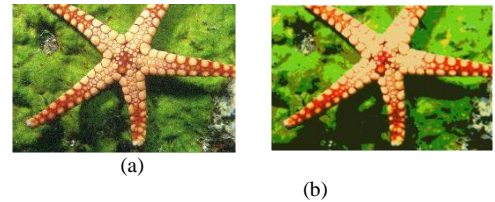

(a) (b)

**Fig. 12:** a) A 30% Gaussian Noise ridden color Image b) Segmented color Image using KWSFCM

### VIII. APPLICATION TO SAR AND MEDICAL IMAGES

Synthetic Aperture Radar (SAR) images are mainly used in remote sensing and mapping of surface lines of earth and other planets. Moreover, SAR images are used in contour detection and in the demarcation process of unknown coastline and terrain. One of the main characteristic of SAR images is that they are prone to speckle noise. Speckle, a multiplicative noise, manifests itself in as apparently random placement of pixels which are conspicuously bright or dark. This noise mainly varies according to the area reflectivity of the test image. High reflectivity introduces high intensity speckle noise where low reflectivity shows low intensity speckle. Two speckle noise-ridden test SAR images have taken into account where both consist of coastlines, contours, distinguishing linear structures as can be seen from Fig. 13. In case of Magnetic Resonance Imaging, estimating Gaussian noise as the main contributing noise distribution would be an underestimation. Magnetic Resonance Noise mainly obeys a general form of *Rician* Distribution, sometimes also the *Rayleigh* distribution, which originates from the static magnetic field used in the imaging process and depends on the sample image size. Fig. 14 shows an MRI image and the competing segmentation maps.

The segmented images in Fig. 13(b)-(g) show the segmentation results for SAR for all the algorithms. KWSFCM shows perfect detection of contour lines and edges of linear structures even when heavy speckles were present along with varying reflectivity, which is evident from Fig. 13(g). In case of MRI images, a close look at Fig. 14(b)-(f) shows that the segmentation results using existing methods fail to preserve the pertinent image structures whereas Fig. 14(g), as obtained by our method, contains perfectly demarcated blood vessels and contours which were ridden with noise in the original noisy image. It is to be noted that 2-level segmentation has been done on the MR image. Also for a quantitative study, the entropy measures for all competing algorithms are tabulated in Table III and our proposed method achieves lowest entropy as can be seen from the values in Table III.

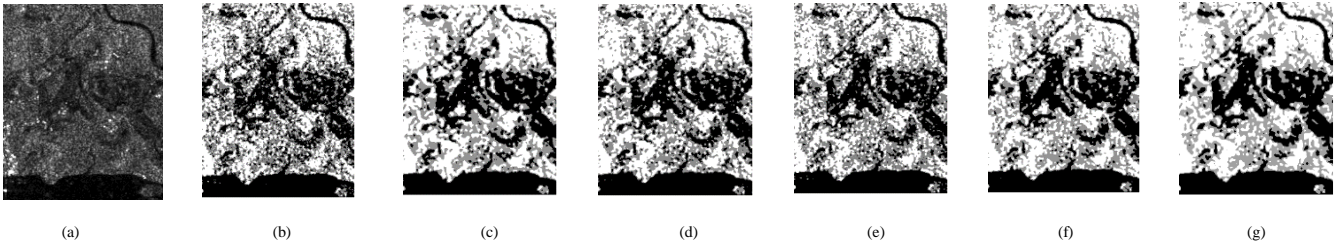

(a)   (b)   (c)   (d)   (e)   (f)   (g)

**Fig. 13**: a) SAR1 image b) NNCut c) FLICM d) RFLICM e) WFLICM f) KWFLICM g) KWSFCM

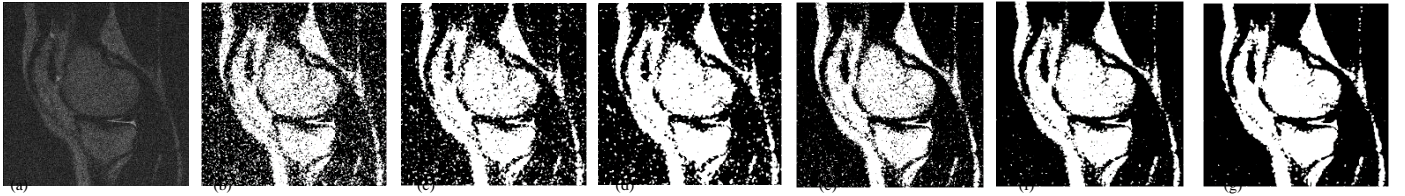

(a)   (b)   (c)   (d)   (e)   (f)   (g)

**Fig. 13**: a) SAR1 image b) NNCut c) FLICM d) RFLICM e) WFLICM f) KWFLICM g) KWSFCM

**Table III.** Entropy measure for SAR and MR images

| Image with noise | Metric | NN Cut | FLICM | RFLICM | WFLICM | KWFLICM | Proposed method |
|---|---|---|---|---|---|---|---|
| SAR1 (Speckle noise) | $H_r(L)$ | 1.0112 | 0.9923 | 0.9848 | 0.9730 | 0.9629 | 0.9596 |
|  | $H_l(L)$ | 0.4778 | 0.4953 | 0.4810 | 0.4862 | 0.4729 | 0.4669 |
|  | E | 1.5890 | 1.4876 | 1.4658 | 1.4592 | 1.4358 | **1.4265** |
| SAR2 (Speckle noise) | $H_r(L)$ | 1.6533 | 1.6397 | 1.6108 | 1.6065 | 1.5878 | 1.5686 |
|  | $H_l(L)$ | 0.3734 | 0.3808 | 0.3898 | 0.3867 | 0.3839 | 0.3790 |
|  | E | 2.0267 | 2.0205 | 2.0006 | 1.9932 | 1.9717 | **1.9476** |
| MR1 (Rician noise) | $H_r(L)$ | 1.2362 | 1.2123 | 1.2021 | 1.1996 | 1.1821 | 1.1727 |
|  | $H_l(L)$ | 0.3460 | 0.3644 | 0.3600 | 0.3437 | 0.3580 | 0.3585 |
|  | E | 1.5822 | 1.5767 | 1.5621 | 1.5433 | 1.5401 | **1.5312** |
| MR2 (Rician noise) | $H_r(L)$ | 0.9102 | 0.8913 | 0.8844 | 0.8710 | 0.8632 | 0.8598 |
|  | $H_l(L)$ | 0.3786 | 0.3965 | 0.3807 | 0.3878 | 0.3723 | 0.3671 |
|  | E | 1.2888 | 1.2878 | 1.2651 | 1.2588 | 1.2355 | **1.2269** |

## IX. A BRIEF LOOK AT THE COMPUTATIONAL TIME OF THE COMPETING ALGORITHMS

KWSFCM accurately segments a noise-ridden image while removing noise and still maintaining proper edge and contour information. The computational time was evaluated after averaging through 25 runs for 20 test images, all of sizes 481x321, taken from the BSDS-500. For the results provided in Table IV, the experiments are carried out on a PC with a second generation core i7 processor running at 2.1 GHZ and having 4 GB RAM. The operating system is Windows 7 home basic and the compiler is MATLAB 7.14.0.139.

**Table IV:** Average computational time per image taken by the competing algorithms

| Competing algorithms | Mean computational time in seconds |
|---|---|
| NNCUT | 3.064 |
| FLICM | 512.613 |
| RFLICM | 406.212 |
| WFLICM | 612.321 |
| KWFLICM | 649.224 |
| KWSFCM | **383.844** |

As is evident from the values in Table IV, NNCut algorithm requires minimum computational time since it involves spectral grouping and does not work on individual windows. However, the NNCut algorithm is not noise immune and hence does not serve the purpose of a good noisy image segmentation. KWSFCM achieves lesser computational time than the other algorithms which also incorporate spatial information into account. Fig. 16 shows the variation of computational time when the image size of the synthetic image, given in Fig. 15(a), is varied from 100x100 to 600x600. The image was Salt & Pepper noise ridden as shown in Fig. 15(b) and Fig. 15(c) shows segmented image using KWSFCM.

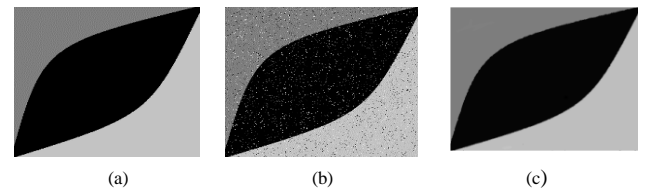

(a)   (b)   (c)

**Fig. 15**: a) Synthetic Image of size 100x100 b) Salt & Pepper noise ridden Synthetic Image c) Segmented image using KWSFCM

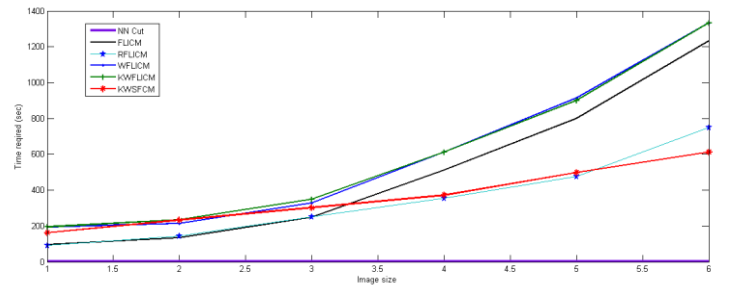

**Fig. 16**: Variation of computational time versus Image size for all competing algorithms

# X. CONCLUSION AND FUTURE WORK

KWSFCM serves as a robust image segmentation algorithm that accurately removes noise in case of noisy images and still maintains the structural characteristics of the image. The proposed algorithm shows appreciable performance for all sorts of noises. The method incorporates weighted SUSAN based fuzzy damping coefficients that increase the contribution of the nucleus with decreasing homogeneity in its neighborhood. However, the parameter $\sigma$ of the kernel has not been made adaptive since a variation of $\sigma$ from 5 to 5000 did not reflect any appreciable change in the performance of the algorithm. Future research work may include:-

a) Investing of other Kernel functions which would require adaptive parameter tuning in pertinence with the test image to be segmented.

b) Extension of such a spatially and circularly weighted SUSAN area algorithm to biomedical image processing for the detection of outliers and other inhomogeneities like fractures and micro-aneurysms.

REFERENCES:

[1] D.L. Pham, and J.L. Prince, "An adaptive fuzzy C-means algorithm for image segmentation in the presence of intensity inhomogeneities," *Pattern RecognitionLetters*, vol. 20, pp.57-68, 1999.

[2] Gath I. and Geva A.B., "Unsupervised Optimal Fuzzy Clustering", *IEEE Transactions on Pattern Analysis and Machine Intelligence*, 11(7), 773–781, 1989.

[3] X. Yin, S. Chen, E. Hu, and D. Zhang, "Semi-supervised clustering with metric learning: An adaptive kernel method," *Pattern Recognit.*, vol. 43, no. 4, pp. 1320–1333, Apr. 2010.

[4] L. Zhu, F. Chung, and S. Wang, "Generalized fuzzy C-means clustering algorithms with improved fuzzy partitions," *IEEE Trans. Syst., Man, Cybern., B, Cybern.*, vol. 39, no. 3, pp. 578–591, Jun. 2009.

[5] S. Tan and N. A. M. Isa, "Color image segmentation using histogram thresholding fuzzy C-means hybrid approach," *Pattern Recognit.*, vol. 44, no. 1, pp. 1–15, 2011

[6] Y.A Toliasand S.M. Panas, "On applying spatial constraints in fuzzy image clustering using a fuzzy rule-based system," *IEEE Signal Processing Letters*, vol. 5, pp.245-247, 1998.

[7] W. Duch, R. Adamczak and K. Grabczewski, "A New Methodology of Extraction, Optimization and Application of Crisp and Fuzzy Logical Rules," *IEEE Transactions on Neural Networks*, vol. 12, pp. 277–306, 2001.

[8] J. Dunn, "A fuzzy relative of the ISODATA process and its use in detecting compact well-separated clusters," *J. Cybern.*, vol. 3, no. 3, pp. 32–57, 1974.

[9] J. Bezdek, *Pattern Recognition with Fuzzy Objective Function Algorithms*. New York: Plenum, 1981.

[10] L. Szilagyi, Z. Benyo, S. Szilagyii, and H. Adam, "MR brain image segmentation using an enhanced fuzzy C-means algorithm," in *Proc. 25th Annu. Int. Conf. IEEE EMBS*, Nov. 2003, pp. 17–21.

[11] W. Cai, S. Chen, and D. Zhang, "Fast and robust fuzzy C-means clustering algorithms incorporating local information for image segmentation," *Pattern Recognit.*, vol. 40, no. 3, pp. 825–838, Mar.2007.

[12] S. Krinidis and V. Chatzis, "A robust fuzzy local information C-means clustering algorithm," *IEEE Trans. Image Process.*, vol. 19, no. 5, pp 1328–1337, May 2010.

[13] M. Gong, Z. Zhou, and J. Ma, "Change detection in synthetic aperture radar images based on image fusion and fuzzy clustering," *IEEE Trans. Image Process.*, vol. 21, no. 4, pp. 2141–2151, Apr. 2012.

[14] N. Cristianini and J. S. Taylor, *An Introduction to SVM's and Other Kernel-Based Learning Methods*. Cambridge, U.K.: Cambridge Univ. Press, 2000.

[15] X. Yang and G. Zhang, "A kernel fuzzy C-means clustering-based fuzzy support vector machine algorithm for classification problems with outliers or noises," *IEEE Trans. Fuzzy Syst.*, vol. 19, no. 1, pp. 105–115, Feb. 2011.

[16] V. Roth and V. Steinhage, "Nonlinear discriminant analysis using kernel functions," in *Advances in Neural Information Processing Systems 12*, S. A Solla, T. K. Leen, and K.-R. Muller, Eds. Cambridge, MA: MIT Press, 2000, pp. 568–574.

[17] B. Scholkopf, A. J. Smola, and K. R. Muller, "Nonlinear component analysis as a kernel eigenvalue problem," *Neural Comput.*, vol. 10, no. 5, pp. 1299–1319, 1998.

[18] S. Chen and D. Zhang, "Robust image segmentation using FCM with spatial constraints based on kernel-induced distance measure",*IEEE Trans. Sys., Man And Cybern..,Part B,*Vol. 34,no. 4,pp 1907-1916, 2004.

[19] M. Gong, Y. Liang, J. Shi, W. Ma, and J. Ma, 'Fuzzy C-Means Clustering With Local Information and Kernel Metric for Image Segmentation'*IEEE Trans. Image Processing* Vol.22, No 2, pp 573-584, Feb 2013.

[20] G. Fijavz, M. Juvan, B. Mohar, and R. Skrekovski, "Circular colorings of planar graphs with prescribed girth", manuscript (2001).

[21] S. M. Smith and J. M. Brady, "SUSAN: A new approach to low level image processin*g*", *International Journal of Computer Vision*, Vol. 23, Issue1, pp.45-78, 1987.

[22] M.Hess and G.Martinez," Facial feature detection based on the smallest univalue segment assimilating nucleus (susan) algorithm", in *Picture Coding Symposium*, San Francisco, California, Dec. 2004.

[23] D. Van De Ville , M. Nachtegael , D. Van der Weken , E. E. Kerre and W. Philips "Noise reduction by fuzzy image filtering", *IEEE Trans. Fuzzy Syst.*, vol. 11, no. 8, pp.429 -436, 2003.

[24] M. G. Choi , J. H. Jung and J. W. Jeon "No-reference image quality assessment using blur and noise", *Proc. World Acad. Sci., Eng. Technol.*, vol. 38, pp.163 -167 2009.

[25] A. Lapini, T. Bianchi, F. Argenti, L. Alparone, "Blind Speckle Decorrelation for SAR Image Despeckling",,*IEEE Trans. Geosc. And Remote Sensing*,Vol 52, Issue 2,pp 1044-1058, Dec 2013.

[26] S. Solbo and T. Eltoft, "Homomorphic Wavelet-based Statistical Despeckling of SAR images, *IEEE Trans. Geosc. And Remote Sensing*,Vol. 42,no 4,pp. 711-721,April 2004.

[27] J.C. Bezdek, L.O. Hall, and L.P. Clarke, "Review of MR image segmentation techniques using pattern recognition," *Medical Physics*, vol.20, pp.1033-1048, 1993.

[28] D.L. Pham, C.Y. Xu, J.L. Prince, "A survey of current methods in medical image segmentation," *Annual Review of Biomedical Engineering*, vol. 2, pp. 315-337, 2000.

[29] V. Boskovitz and H. Guterman, "An adaptive eneuro-fuzzy system for automatic Image segmentation and edge detection", *IEEE Transactions on Fuzzy Systems*, Vol. 10, Issue 2, pp. 247-261, 2002.

[30] I. Bloch, "Fuzzy sets in image processing", *Proceedings of ACM Symposium on Applied Computing*, New York, USA, March 6-8, pp. 175 –179, 1994.

[31] J. C. Bezdek, R. Chandrasekhar and Y. Attikiouzel, "A geometric approach to edge detection", *IEEE Transactions on Fuzzy Systems*, Vol. 6, Issue l, pp. 52- 75, 1998.

[32] C. Fowlkes, S. Belongie, F. Chung, and J. Malik, "Spectral grouping using the Nystrom method," *IEEE Trans. Pattern Anal. Mach. Intell.*, vol. 26, no. 2, pp. 1–12, Feb. 2004.

[33] C. Li, R. Huang, Z. Ding, J. C. Gatenby, D. N. Metaxas, and J. C. Gore, "A level set method for image segmentation in the presence of intensity inhomogeneities with application to MRI," *IEEE Trans. Image Process.*, vol. 20, no. 7, pp. 2007–2016, Jul. 2011.

[34] H. Zhang, J. Fritts, and S. Goldman, "An entropy-based objective evaluation method for image segmentation," *Proc. SPIE, Storage Retrieval Methods Appl. Multimedia*, vol. 5307, pp. 38–49, Jan. 2004.